\documentclass{article}
\usepackage{ijcai26}

\usepackage{times}
\usepackage{soul}
\usepackage{url}
\usepackage[hidelinks]{hyperref}
\usepackage[utf8]{inputenc}
\usepackage[small]{caption}
\usepackage{graphicx}
\usepackage{amsmath}
\usepackage{amsthm}
\usepackage{booktabs}
\usepackage{algorithm}
\usepackage{algorithmic}
\usepackage[switch]{lineno}

\usepackage{listings}
\usepackage{multirow}
\usepackage{xcolor}
\usepackage{enumitem}
\usepackage[acronym]{glossaries}

\graphicspath{{./images/}{images/}}

\DeclareMathOperator*{\argmin}{argmin}

\newcommand{\coco}{\texttt{coco}}
\newcommand{\celeb}{\texttt{celeb}}
\newcommand{\pa}{PixelArena}

\newcommand{\eg}{\emph{e.g.}}
\newcommand{\ie}{\emph{i.e.}}

\makeglossaries

\newacronym{t2i}{T2I}{text to image}
\newacronym{omm}{OMM}{omni-modal model}
\newacronym{vlm}{VLM}{visual language model}
\newacronym{cot}{CoT}{chain of thoughts}
\newacronym{ppvi}{PPVI}{Pixel-Precision Visual Intelligence}
\newacronym{ss}{SS}{semantic segmentation}

\newacronym{geminipro}{\texttt{gmn3}}{Gemini 3 Pro Image}
\newacronym{gemini}{\texttt{gmn25}}{Gemini 2.5 Flash Image}
\newacronym{gpt}{\texttt{gpti}}{GPT Image 1}
\newacronym{sam3}{\texttt{sam3}}{SAM 3}
\newacronym{emu35}{\texttt{emu35}}{Emu 3.5}
\newacronym{unimoe2}{\texttt{unimoe2}}{Uni-MoE-2}
\newacronym{unimoe2image}{\texttt{unimoe2-image}}{Uni-MoE-2 Image}
\newacronym{unimoe2omni}{\texttt{unimoe2-omni}}{Uni-MoE-2 Omni}
\newacronym{segface}{\texttt{segface}}{SegFace}
\newacronym{oneformer}{\texttt{1former}}{OneFormer}

\title{PixelArena: A Benchmark for Pixel-Precision Visual Intelligence}

\author{
Feng Liang$^{1}$\thanks{These authors contributed equally to this work.}
\and
Sizhe Cheng$^{1}$\footnotemark[1]
\and
Chenqi Yi$^{1}$
\And
Yong Wang$^{1}$\thanks{Corresponding author}\\
\affiliations
$^1$Nanyang Technological University, 50 Nanyang Avenue, Singapore\\
\emails
\{feng011, sizhe003, chenqi001\}@e.ntu.edu.sg,
yong.wang@ntu.edu.sg
}

\begin{document}

\maketitle

\begin{abstract}
Omni-modal models that have multimodal input and output are emerging. However, benchmarking their multimodal generation, especially in image generation, is challenging due to the subtleties of human preferences and model biases. Many image generation benchmarks focus on aesthetics instead of the fine-grained generation capabilities of these models, failing to evaluate their visual intelligence with objective metrics. In \pa{}, we propose using semantic segmentation tasks to objectively examine their fine-grained generative intelligence with pixel precision. 
With our benchmark and experiments, we find the latest \acrlong{geminipro} has emergent image generation capabilities that generate semantic masks with high fidelity under zero-shot settings, showcasing visual intelligence unseen before and true generalization in new image generation tasks. 
We further investigate its results, compare them qualitatively and quantitatively with those of other models, and present failure cases. The findings not only signal exciting progress in the field but also provide insights into future research related to dataset development, omni-modal model development, and the design of metrics.
\end{abstract}
\section{Introduction}
\label{sec:intro}

Since the release of GPT-4o~\cite{gpt4o} in 2024, \glspl{omm}, which have multiple input and output modalities (\eg, text, images, and audio), have been a focus of research. Numerous \glspl{omm} have been developed (\eg, Emu series~\cite{emu1,emu2,emu3,emu35}, Gemini series~\cite{gemini3proimage,gemini25flashimage}). They can generate images based on prompts that include both text and images. This capability is highly malleable, enabling flexible in-context learning and powerful, convenient, conversational image generation. However, as much focus has been placed on image quality and aesthetics, few have quantitatively examined the precision and generalizability of the image generation capabilities of these models, nor have they examined the limitations of visual reasoning and perception of these models during image generation.

To address the aforementioned issues, in \pa{}, we propose using pixel-level tasks, the ones in \gls{ss}, to examine \glspl{omm}' fine-grained control capability (\eg, painting individual pixels with precise colors) and their generalizability (\ie, generalizing to new pixel-level tasks) in image generation, which we term \gls{ppvi}. By further examining \glspl{omm}' reasoning process, in \pa{}, we can also unveil their limitations in visual reasoning and perception. Specifically, we ask models to perform \gls{ss} tasks on subsets of CelebAMask-HQ~\cite{CelebAMask-HQ} and COCO~\cite{coco-dataset} as two examples. This allows us to use objective metrics (\eg, F1 Score, mIoU, and Dice) to measure fine-grained generative capability. We select strong \glspl{omm} that were released within the last six months, including \acrlong{geminipro}~\cite{gemini3proimage}, \acrlong{gemini}~\cite{gemini25flashimage}, \acrlong{gpt}~\cite{gpt4o}, \acrlong{emu35}~\cite{emu35}, and \acrlong{unimoe2}~\cite{unimoe2}. In our experiments, we measure quantitative results to evaluate the performance of \glspl{omm} on the datasets. We also develop a graphical interface\footnote{\url{https://pixelarena.reify.ing}} to qualitatively examine the results. With these results, we find that \acrlong{geminipro} represents a significant leap in this front, compared to other models. With the quantitative results, we also show that \acrlong{geminipro} truly generalizes to new image generation tasks. We also present interesting failure cases and analyze their implications.

In summary, our contributions are:
\begin{enumerate}[leftmargin=2.5em]
    \item We propose a benchmark, \pa{}, in which pixel-level tasks (\ie, \gls{ss} tasks) are used to quantitatively measure \glspl{omm}' \gls{ppvi}, including fine-grained control capability and generalizability of their image generation capabilities.
    \item We task \glspl{omm} with face parsing using the CelebAMask-HQ~\cite{CelebAMask-HQ} dataset, revealing surprising emergent zero-shot capabilities in \acrlong{geminipro}~\cite{gemini3proimage}. We also perform experiments to examine potential data contamination in this model, showing that the model does not memorize the answers (\ie, reference masks) but truly understands this image generation task. We also present the \gls{ss} results on a significantly more challenging dataset, COCO~\cite{coco-dataset}, showing that \acrlong{geminipro} still has reasonable performance and generalization.
    \item We conduct both qualitative and quantitative analyses of the results, including failure cases, hinting at more future directions in dataset development, \gls{omm} research, and design of metrics.
\end{enumerate}

\section{Related Work}
\label{sec:related_work}

\subsection{Semantic Segmentation and State-of-the-Art Models}

In computer vision research, various segmentation datasets have been developed, such as COCO~\cite{coco-dataset}, a large-scale benchmark containing object-centric images with pixel-level annotations for diverse everyday scenes; FSS-1000~\cite{fss-1000}, a few-shot segmentation dataset featuring 1,000 object categories with only a single annotated example per class; SA-CO~\cite{sam3}, which extends segment-anything-style annotation to concept-driven segmentation tasks.

In this research, we use face parsing, an \gls{ss} task, with the CelebAMask-HQ~\cite{CelebAMask-HQ} dataset as an example, a high-quality facial image dataset that provides detailed pixel-level annotations for 18 distinct facial components across 30,000 celebrity images. Another example we present is general \gls{ss} on the panoptic segmentation dataset of COCO~\cite{coco-dataset}.

Various models have been proposed to push the state of the art on CelebAMask-HQ~\cite{CelebAMask-HQ} and COCO~\cite{coco-dataset}. The latest SegFace~\cite{segface} improves the state of the art on CelebAMask-HQ by explicitly addressing long-tail facial components through a balanced segmentation framework. OneFormer~\cite{oneformer} and Mask2Former~\cite{mask2former} are the state-of-the-art models on the panoptic segmentation dataset of COCO~\cite{coco-dataset}. They are capable of performing universal image segmentation (\ie, \gls{ss}, instance segmentation, and panoptic segmentation).

Note that in \pa{}, we do not intend to use \glspl{omm} to compete with these state-of-the-art specialized models that are specifically designed and trained for \gls{ss} on specific datasets; instead, we probe the emergent generative capabilities and generalizability of these generalist models (\ie, \glspl{omm}). In our experiments, we task \glspl{omm} to generate masks with published weights or public APIs and no further training.

Another line of research focuses on integrating a \gls{vlm} that has text and image input but \textit{text-only output} with a segmentation model. RAS~\cite{ras} enhances segmentation models by integrating a mask-centric large \gls{vlm} that selects relevant mask groups from a pool of candidates based on vision-language prompts, enabling flexible and precise mask grouping. SAM4MLLM~\cite{sam4mllm} trains a \gls{vlm} to output prompts (bounding boxes and points) to guide SAM~\cite{sam1} in generating accurate segmentation masks, thus combining language understanding with pixel-level mask generation. The large \gls{vlm} for remote sensing images~\cite{Liu31122025} uses a language model to interpret open-vocabulary queries and conditions a segmentation decoder to produce class-specific masks, enabling flexible, high-resolution \gls{ss} of unseen categories. All of them integrate a \gls{vlm} and a segmentation model with text or latent vectors as intermediate representations. However, in \pa{}, we use the original \glspl{omm} \textit{without} any tools, model integration, or finetuning. 

Another noteworthy work is SAM~3\cite{sam3}. In this work, SAM Agent is proposed, which is similar to SAM4MLLM~\cite{sam4mllm}. It also presents preliminary results generated by \acrlong{gemini} in object detection tasks with ODinW13~\cite{ODinW13} and RF-100VL~\cite{RF-100VL} using prompts and few-shot learning. This method is similar to ours, but its output is bounding box coordinates, whereas ours are mask images. In contrast to testing visual perception by generating text (\ie, coordinates), we test the finer-grained generative capabilities of \glspl{omm} at the pixel level by generating images (\ie, masks). Instead of providing examples, we give high-level instructions to the models to generate masks, forcing them to perform our tasks in \textit{zero-shot} settings.

\subsection{Image Generation Benchmarks and Metrics}

Most of the benchmarks~\cite{lee2023holisticevaluationtexttoimagemodels,t2ibench,hu2023tifaaccurateinterpretabletexttoimage,yu2022scalingautoregressivemodelscontentrich,saharia2022photorealistictexttoimagediffusionmodels,hu2024ellaequipdiffusionmodels,ku2024imagenhubstandardizingevaluationconditional,zhang2024magicbrushmanuallyannotateddataset,sheynin2023emueditpreciseimage,ye2025imgedit,jayasumana2024rethinkingfidbetterevaluation} for text-to-image generation and image editing focus on evaluating generated natural images rather than the masks that we use. However, due to their diversity and complexity, natural images are difficult to evaluate, and the evaluation metrics are often subject to implicit human preferences or model biases. 

For example, in HEIM~\cite{lee2023holisticevaluationtexttoimagemodels}, the metrics used are CLIP score, FID, the score from a LAION aesthetics predictor, human evaluation score, and VQA-based scores. However, the CLIP score, FID, aesthetics score and VQA-based scores may be biased by the models used, while human evaluation is fundamentally based on implicit preferences. In ImgEdit~\cite{ye2025imgedit}, GPT-4o~\cite{gpt4o} is prompted with detailed scoring rubrics based on three dimensions (\ie, instruction adherence, image-editing quality, and detail preservation) to score the generated images. It also incorporates a forensic detector, FakeShield~\cite{xu2024fakeshield}, to compute a fake score for the generated images. FakeShield~\cite{xu2024fakeshield} also uses models for scoring, including GPT-4o~\cite{gpt4o} and fine-tuned models based on SAM~\cite{sam1} and Qwen2.5-VL~\cite{bai2025qwen25vltechnicalreport}. These metrics are fundamentally subjective with respect to human preferences or model biases.

In \pa{}, because we task models to generate masks and evaluate the generated masks, we can use standard objective metrics such as F1 Score and mIoU.
\section{\pa{}}

\subsection{Dataset}

We use the COCO~\cite{coco-dataset} and CelebAMask-HQ~\cite{CelebAMask-HQ} datasets as examples. As these datasets contain thousands of images and \glspl{omm} are computationally demanding, we randomly sampled 150 images and their corresponding masks from each dataset. With sufficient resources, we can conduct experiments on the entire datasets in the future.

For the CelebAMask-HQ~\cite{CelebAMask-HQ} dataset, we first perform random sampling to obtain a small subset. As the reference masks are $512\times512$, while the selected \glspl{omm}~\cite{gemini25flashimage,gemini3proimage,gpt4o,unimoe2,emu35} natively support image generation with resolutions larger than $512\times512$ (\eg, $720\times720$ and $1024\times1024$), we upsample the reference masks using nearest neighbors to $1024\times1024$. We also upsample the generated masks using nearest neighbors to $1024\times1024$.

For the COCO~\cite{coco-dataset} dataset, we perform the same sampling process on its panoptic segmentation dataset. We convert the panoptic masks into \gls{ss} masks using its official toolkit\footnote{https://github.com/cocodataset/panopticapi}. As the resolutions of the images and reference masks in the dataset are not fixed, we center-crop them based on the shortest dimension to obtain square ones. Similarly, we upsample the reference masks and generated masks to $1024\times1024$ using nearest neighbors. We evaluate the metrics (\ie, F1 Score, mIoU, and Dice) using the processed reference masks and predicted masks generated from the processed images, ensuring the fairness of the evaluation.

In the following sections, we refer to these two subsets as the \celeb{} and \coco{} datasets, respectively. We use three metrics (\ie, F1 Score, mIoU, and Dice) to evaluate the performance of the selected \glspl{omm} on the two datasets.

\subsection{Models and Mask Generation}

For different models, we use different methods to generate valid segmentation masks. For the sake of brevity, we will refer to the selected models by their short code names (\eg, \acrshort{geminipro}) in the following sections.

\begin{figure}
    \centering
    \includegraphics[width=0.75\linewidth]{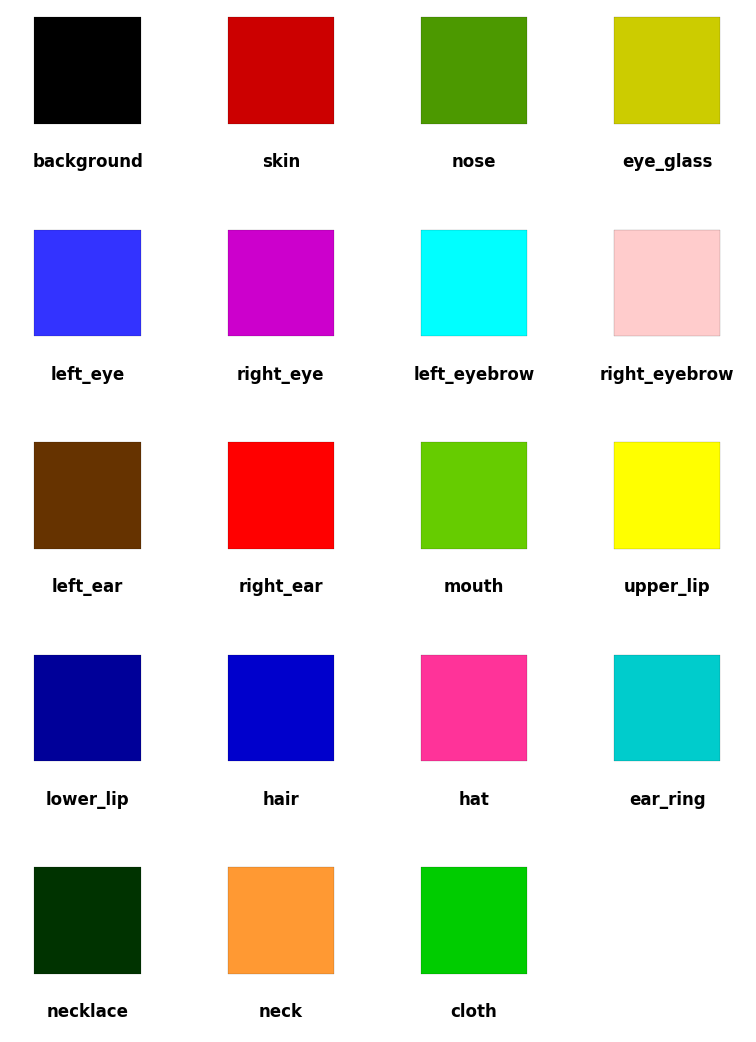}
    \caption{Palette of the standard color encodings from CelebAMask-HQ~\protect\cite{CelebAMask-HQ}.}
    \label{fig:label_palette}
\end{figure}

\begin{lstlisting}[
    float,
    captionpos=b,
    caption={Prompt Template for our \celeb{} experiments. We omit the rest of color codings here. As there is ambiguity in the left and right in terms of references (\ie, with respect to images or persons), we clarify this in length in the prompt to avoid confusion.},
    label=prompt
]
I want you to do semantic segmentation based on facial features. 
The label encodings are

```
background : [0, 0, 0]
... omitted
```

For your convenience, I've also given you a color palette (the second image) for the label encodings.

Please draw a colorful mask, given the photo (the first image), the color palette and the label encodings. 

Note that for the left and right used by the labels, these are with respect to the person in the image, NOT the image itself, so the left facial features of the person are on the right of the image. Check if you have labeled the features on the left of the image to be the right feature labels.
\end{lstlisting}

\textbf{For \glspl{omm}:} We select recent models with strong image generation capabilities, including \acrfull{geminipro}~\cite{gemini3proimage}, \acrfull{gemini}~\cite{gemini25flashimage}, \acrfull{gpt}~\cite{gpt4o}, \acrfull{emu35}~\cite{emu35}, and \acrfull{unimoe2}~\cite{unimoe2}. Note that we tested two variants of \acrshort{unimoe2}: \acrfull{unimoe2omni}, the flagship model of the series, and \acrfull{unimoe2image}, the variant finetuned for image generation. As they natively generate images instead of label vectors, we first prompt them to generate images with specified color encodings and then convert the pixels from RGB values to segmentation class labels. The prompts for the two datasets are composed of three parts: an image from the dataset, an image of the color palette of label encodings (Fig.~\ref{fig:label_palette} and Fig.~2 in Suppl.~A ), and a short text (Listing~\ref{prompt}). These prompts provide task specifications and visual grounding for the color encodings, as well as some clarifications. Note that no examples are given in our prompts, meaning the models have to learn \gls{ss} tasks \textit{zero shot}. For the sampling parameters of the \glspl{omm}, refer to Table~1 in Suppl.~B. Given the mask images generated by \glspl{omm}, we compare the RGB value of each pixel with the color encodings for the labels, selecting the nearest color and label. It is formulated as Eq.~\ref{eq:1}, where $\vec{e_i}$ is the RGB color vector of a label with index $i$, and $\vec{p}$ is the color vector of a pixel.

\begin{equation} \label{eq:1}
i = \argmin_{i}\ (\vec{e_i} \cdot \vec{p})
\end{equation}

\textbf{For SAM 3:} \acrfull{sam3}~\cite{sam3} accepts text as the prompt for mask generation. We prompt \acrshort{sam3} with the labels of CelebAMask-HQ~\cite{CelebAMask-HQ} one by one and merge the corresponding 19 masks into one final mask. For the label of each pixel in the overlapping areas of these masks, we randomly pick one from the overlapping labels.

\textbf{For specialized computer vision models:} We use the pretrained ConvNext~\cite{convnext} variant of \acrfull{segface}~\cite{segface} as a strong baseline model on \celeb{}; while on \coco{}, we use \acrfull{oneformer}~\cite{oneformer}.

\section{Analysis}

Due to the stochastic components (\eg, token sampling, diffusion module) in \glspl{omm}, the generation of mask images is inherently stochastic. Therefore, we present results from multiple attempts. The number of attempts is $p=[1, 3, 5]$.

\subsection{Qualitative Comparisons}

\begin{figure}
    \centering
    \includegraphics[width=0.92\linewidth]{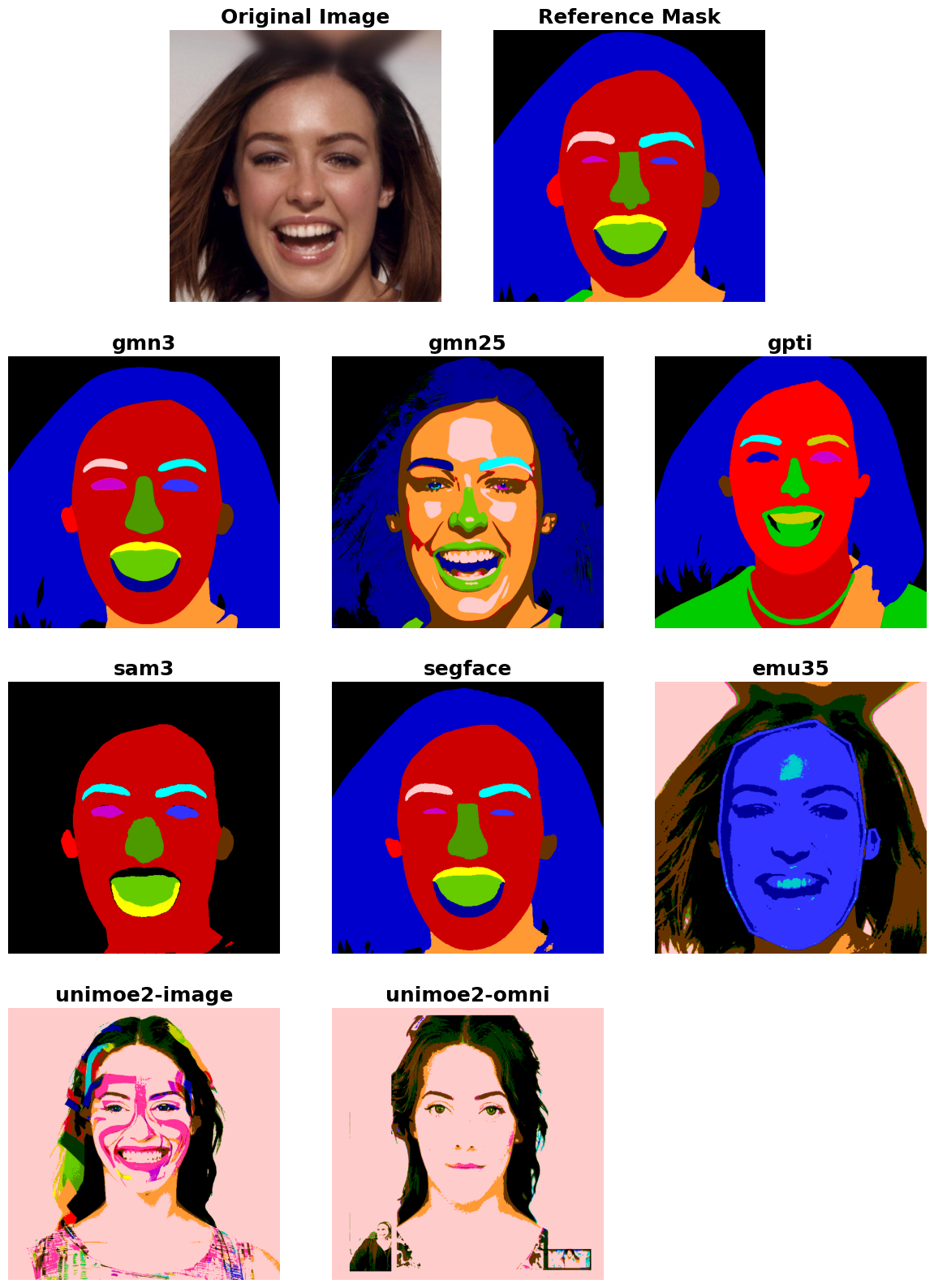}
    \caption{Comparison between the Results of Different Models on \celeb{}. The short code names are shown on the top of images. The results are not cherry-picked.}
    \label{fig:model-compare}
\end{figure}

\begin{figure}
    \centering
    \includegraphics[width=1\linewidth]{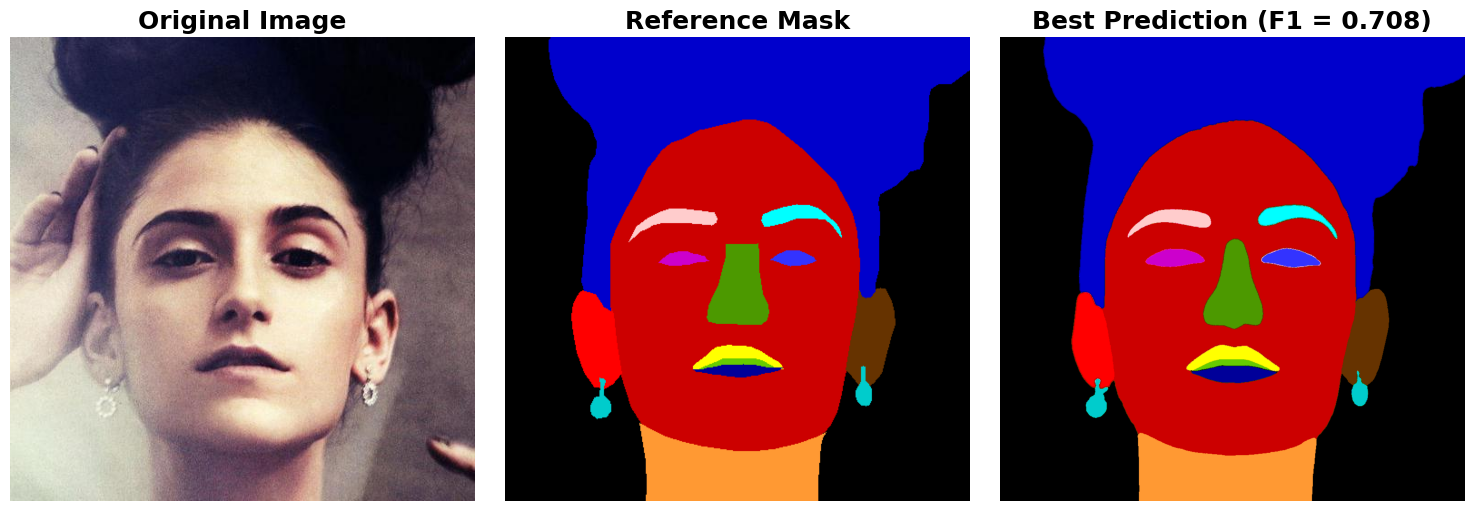}
    \caption{Best prediction across \celeb{} by \acrshort{geminipro} with F1 score $0.708$. The short code names are shown on the top of images.}
    \label{fig:best-gemini-pro}
\end{figure}

\begin{figure}
    \centering
    \includegraphics[width=1\linewidth]{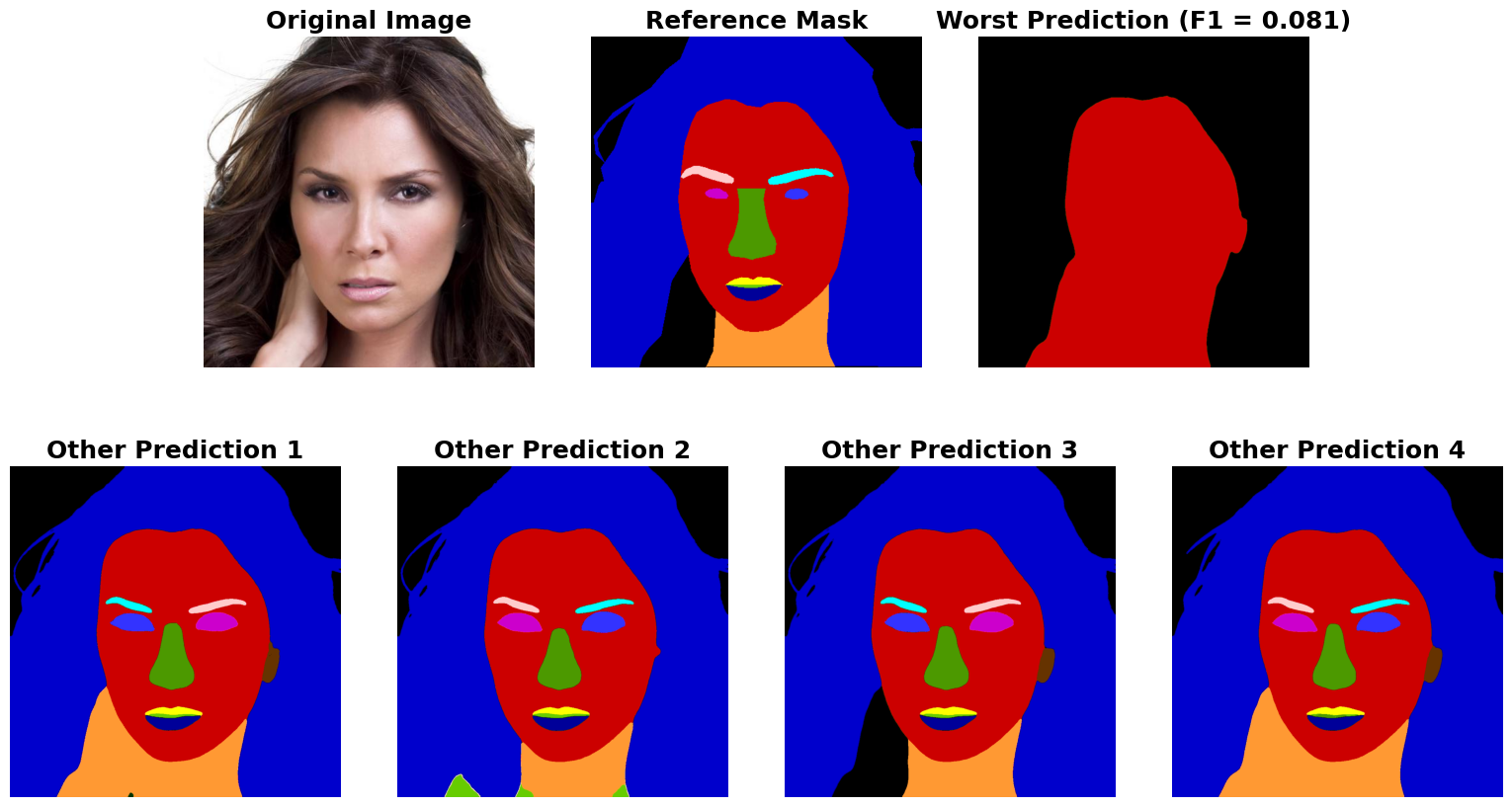}
    \caption{Worst prediction across \celeb{} by \acrshort{geminipro} with F1 score $0.081$ and parallel attempts. Attempt numbers are shown on the top of images.}
    \label{fig:worst-gemini-pro}
\end{figure}

In Fig.~\ref{fig:model-compare}, we present the results of different models on \celeb{} for qualitative comparison. Among all \glspl{omm}, \acrshort{geminipro} is the \textit{only} one that understands the task requirements \textit{and} completes it with high quality. \acrshort{gpt} and \acrshort{gemini}  partially understand the task, but \acrshort{gemini} lacks precise color control or fails to understand the color encodings, while \acrshort{gpt} lacks precise control over the composition of the image and hallucinates the upper body of the person. As for \acrshort{sam3}, it sometimes misses some labels. \acrshort{emu35} and \acrshort{unimoe2} models completely misunderstand the task while presenting different failure patterns. \acrshort{emu35} failed to draw plausible masks, but it could control its image generation process to replicate most features of the original image. In contrast, \acrshort{unimoe2} models could not even draw an image similar to the original image, which may be due to its vision system failing to capture the original image, failing to propagate the visual information to its generation module, or failing to control its generation process. 

Such failure modes are common in the results of the respective models, implying a potential misalignment between the vision system and the generation module, or a lack of control over the generation process.

We further investigate the results of \acrshort{geminipro}. We present the best and worst results from the model in Fig.~\ref{fig:best-gemini-pro} and Fig.~\ref{fig:worst-gemini-pro}. The best prediction has a nearly indistinguishable difference from the reference mask, while the worst prediction is dramatically low in quality. However, the other four attempts in Fig.~\ref{fig:worst-gemini-pro} present reasonable results, which suggests that the generation process is not stable or robust.

\begin{figure}
    \centering
    \includegraphics[width=0.8\linewidth]{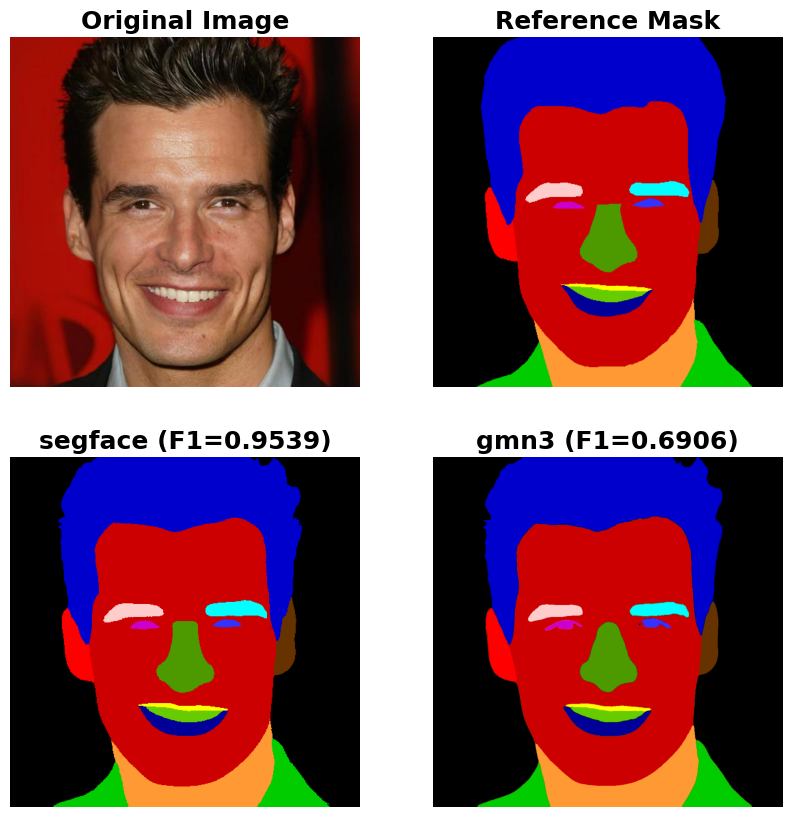}
    \caption{Comparison between the reference mask and masks predicted by two strong models on \celeb{}. Their short code names and F1 scores are on the top of the images. We picked the image on which \acrshort{segface} achieved the highest F1 score $0.9539$ while \acrshort{geminipro} achieved $0.6906$.}
    \label{fig:segface-gemini}
\end{figure}

Although we do not intend to use \glspl{omm} to compete with the specialized computer vision model (\eg, \acrshort{segface}), we present the result in which \acrshort{segface} achieved the highest F1 score $0.9539$, while \acrshort{geminipro} achieved $0.6906$ in Fig.~\ref{fig:segface-gemini}. We find that these two masks are visually very similar, while the scores have a significant gap.

\subsection{Quantitative Results and Examining Data Contamination}

\begin{figure}
    \centering
    \includegraphics[width=1\linewidth]{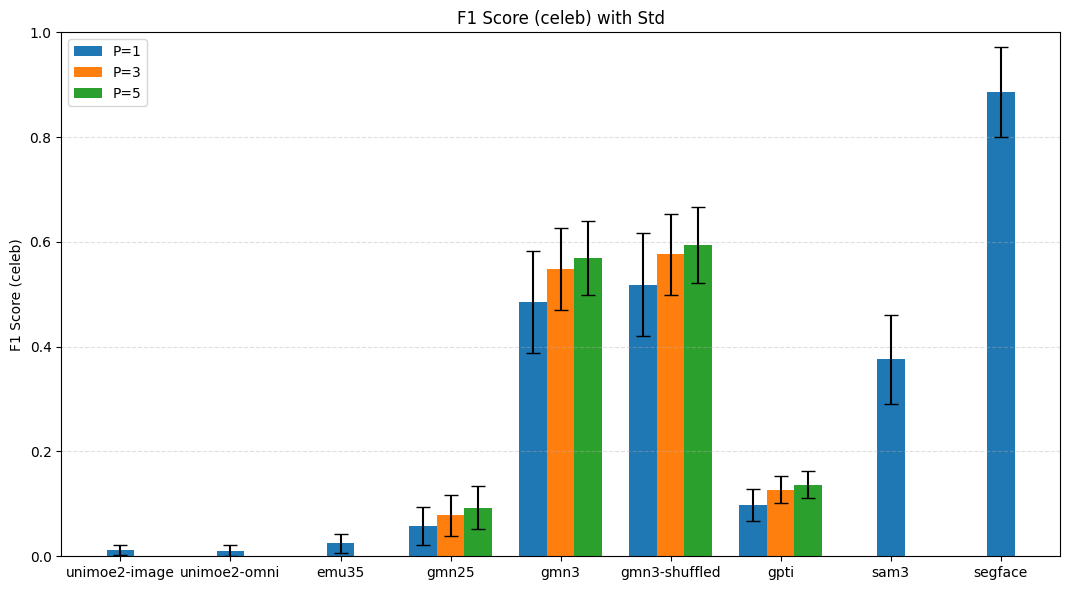}
    \caption{F1 Scores for experiments on \celeb{}. For $p=[3, 5]$, we ask \glspl{omm} to try 3 or 5 times and select the best result in these attempts. As \acrshort{sam3} and \acrshort{segface} contain no stochastic components, we did not run more attempts. Due to their poor performance, we did not run experiments of \acrshort{emu35} and \acrshort{unimoe2} with more attempts.}
    \label{fig:quantatitive}
\end{figure}

We present the F1 scores in Fig.~\ref{fig:quantatitive}. For mIoU and Dice, please refer to Figs.~3 and~4 in Suppl.~C. Aligned with our qualitative analysis, \acrshort{geminipro} achieved the best F1 score, mIoU, and Dice in the selected \glspl{omm}, although it lags behind \acrshort{segface}.

As the results of \acrshort{geminipro} are surprisingly good on \celeb{}, we further check whether data contamination is the cause of such good performance rather than true generalization, since  CelebAMask-HQ~\cite{CelebAMask-HQ} has published all images and masks on the Internet. We shuffle the color encodings (Fig.~1 in Suppl.~A) instead of using the standard encodings in a new experiment.  Noteworthy is that, as shown in Fig.~\ref{fig:quantatitive}, after we shuffled the color encodings, the performance of \acrshort{geminipro} (\texttt{gmn3-shuffled}) did not drop but instead increased by roughly 10\% compared to its original result. This means the model did not memorize the reference masks but truly understood the task, including the required color mapping using arbitrary color encodings.

\subsection{Further Failure Analysis and Pretended Reflections}

After inspecting \glspl{cot} of the results generated by \acrshort{geminipro}, we find that during the image generation process, \acrshort{geminipro} performs a three-step \gls{cot} before presenting the final result: it first considers the task and requirements, then generates a draft image, and finally checks the draft against the requirements and reflects on the result. Such a process seems to imply quality control and iterative refinements. However, as we can see in Fig.~\ref{fig:worst-gemini-pro}, a low quality mask could pass its final check in its \gls{cot}. 

\begin{figure}
    \centering
    \includegraphics[width=1\linewidth]{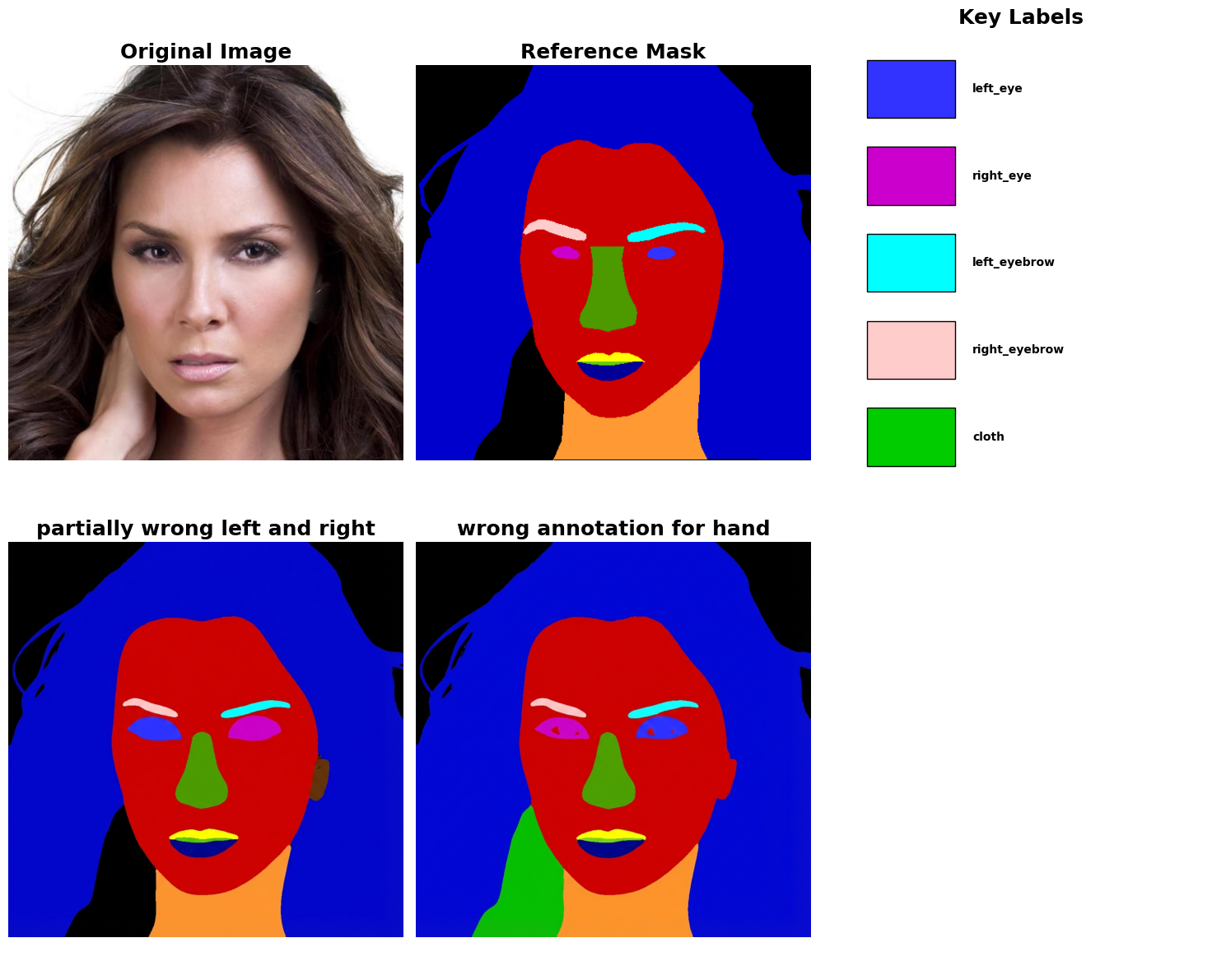}
    \caption{More failure instances. Top Right: The palette of related labels. Bottom left: \acrshort{geminipro} correctly identified the left and right eyebrows while confused about the left and right eyes. Bottom right: It mislabeled the hand as cloth.}
    \label{fig:errors}
\end{figure}

We further investigate such failures with more attempts on the same image. Although we could not reproduce the extreme failure in Fig.~\ref{fig:worst-gemini-pro}, we present two interesting instances in Fig.~\ref{fig:errors}. For the partially incorrect case (bottom left in Fig.~\ref{fig:errors}), \acrshort{geminipro} labeled the right and left eyebrows correctly but mislabeled the eyes, which are in close proximity. In its \gls{cot}, however, it explicitly concluded that ``\textit{I've verified that the segmentation mask strictly adheres to all user-specified constraints. Facial feature delineation, including the critical left/right reversal rule, is accurate...}'' Such pretended quality control also occurred in the second case (bottom right in Fig.~\ref{fig:errors}), in which it misclassified the hand as cloth. In its \gls{cot}, it reflected: ``\textit{... The background, hair, skin, neck, nose, eyebrows, eyes, lips and cloth were all correctly segmented. I'm satisfied that this fulfills all requirements.}''

The reflections of the model are merely pretending, blindly affirming the correctness of the result. This may be a fundamental flaw in its multimodal reasoning.  Such examples may also represent adversarial attacks that target potential flaws and bypass material checking.

\subsection{More experiments on \coco{}}

Even in the failure cases of \acrshort{geminipro} shown in Figs.~\ref{fig:worst-gemini-pro} and~\ref{fig:errors}, most of the results are still reasonably good, which makes us wonder whether face parsing on \celeb{} is too easy for advanced \glspl{omm} such as \acrshort{geminipro}. Therefore, we tested the performance of \acrshort{geminipro} and \acrshort{gemini} and compared them with \acrshort{oneformer} on \coco{}. Compared to \celeb{}, \coco{} is much more challenging, since its number of classes is much larger (144 vs. 19). We present and analyze the best and worst predictions of \acrshort{geminipro} in this section.

\begin{figure}
    \centering
    \includegraphics[width=1\linewidth]{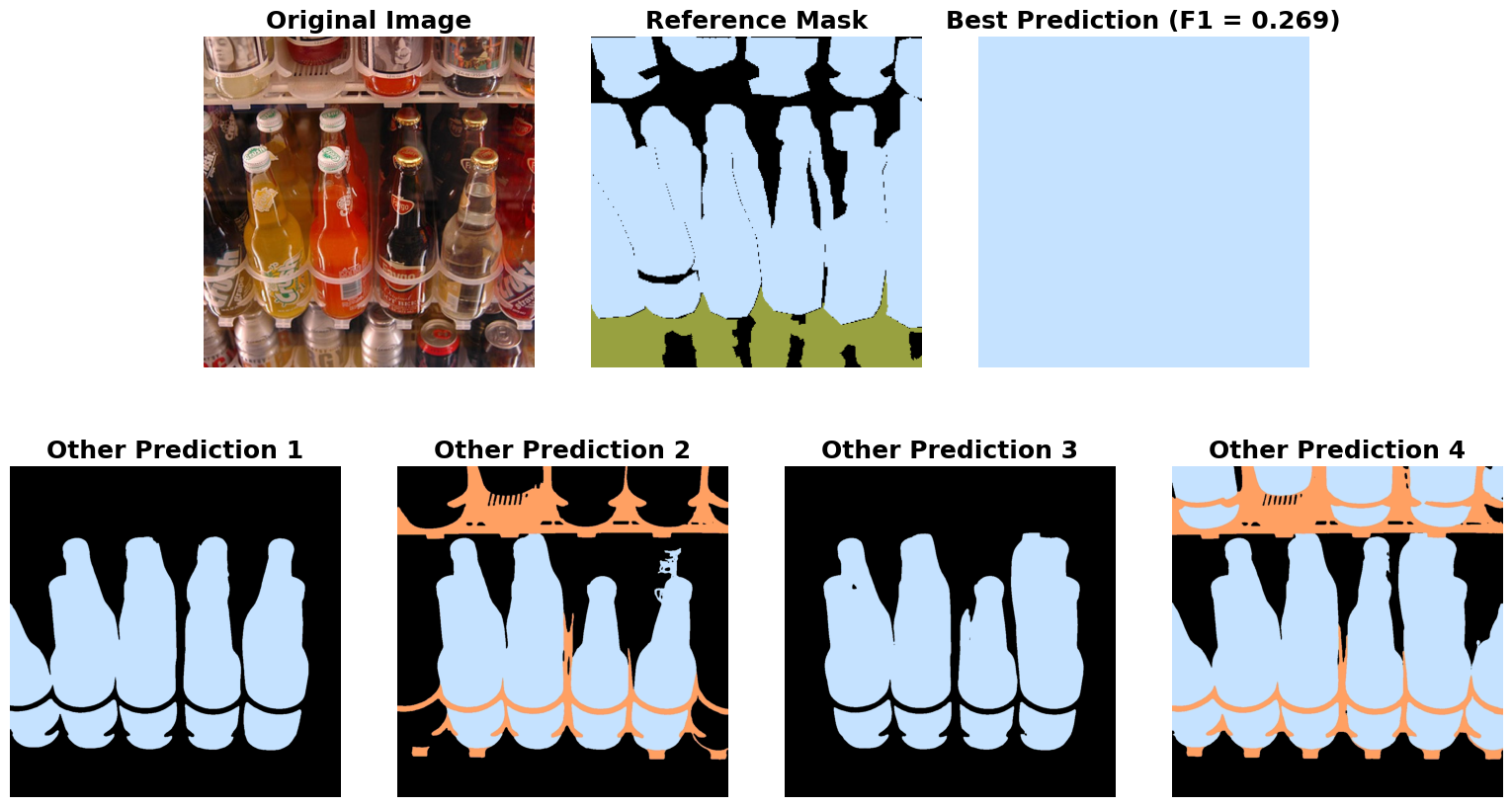}
    \caption{Best prediction (Top Right) of \acrshort{geminipro} on \coco{}, achieving F1 score $0.269$. The bottom row showcases other four attempts.}
    \label{fig:geminipro-best-coco}
\end{figure}

\begin{figure}
    \centering
    \includegraphics[width=1\linewidth]{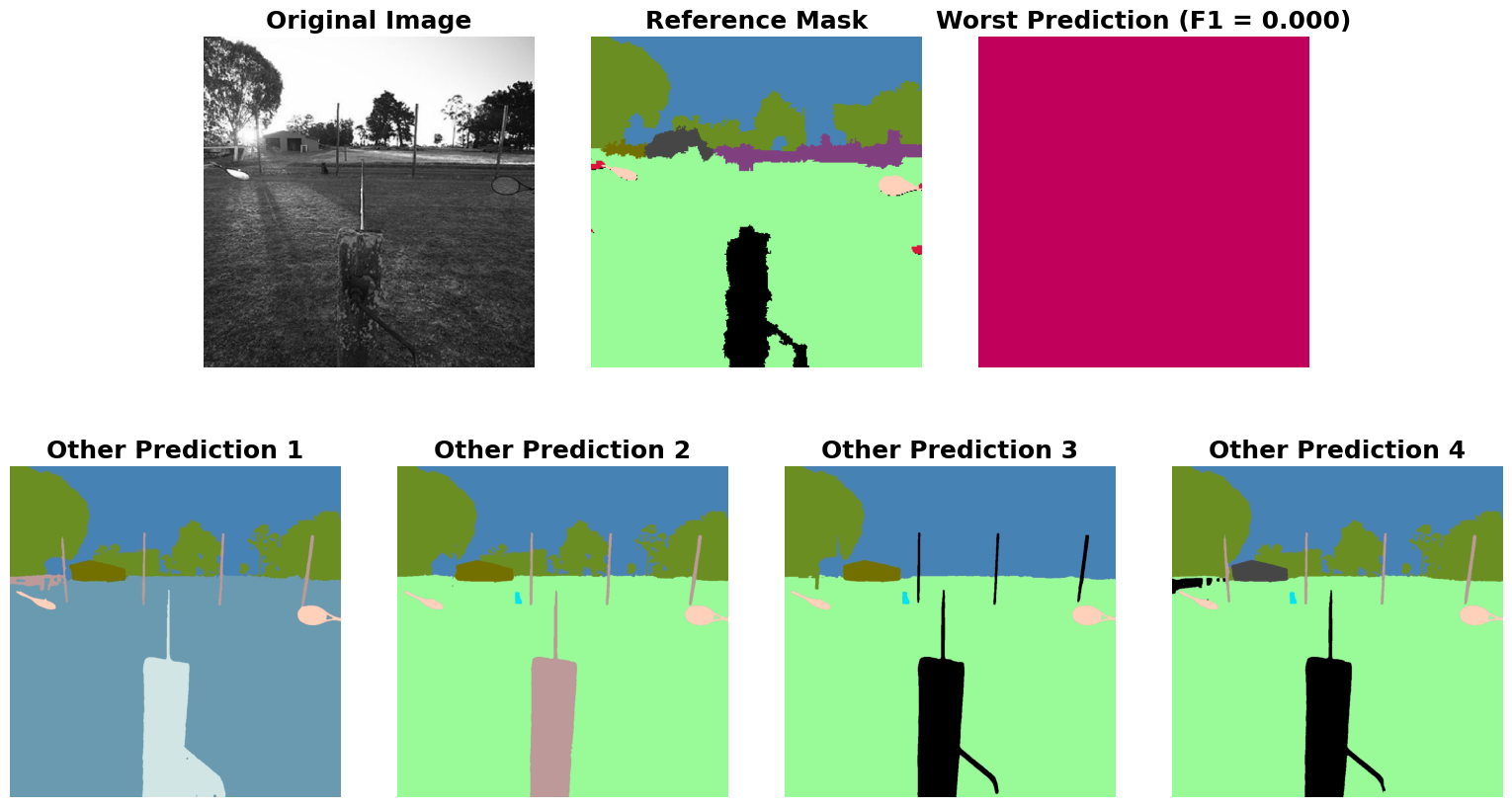}
    \caption{Worst prediction (Top Right) of \acrshort{geminipro} on \coco{}, achieving F1 score $0.0$. The bottom row showcases other four attempts.}
    \label{fig:geminipro-worst-coco}
\end{figure}

The best result (Fig.~\ref{fig:geminipro-best-coco}) is interesting in that \acrshort{geminipro} seems to give up on drawing a detailed mask; yet, the F1 score is higher than those of other attempts, which appear more plausible because, in this example, the area with the ``bottle'' label is dominantly larger than the areas of other labels. Additionally, the issue of pretended reflections persists in this case, as the reflection of the model is:
\begin{quote}
\textit{I'm currently verifying the semantic segmentation mask's consistency with the input image. I've analyzed the color mapping for the ``bottle'' category, expecting a uniform light blue representation. The generated output is a solid light blue square, which accurately reflects the category distribution in the original image. There appear to be \textit{no} other categories present in the output.}
\end{quote}

As seen in the reference mask and the results of other attempts, there are categories other than ``bottle'' in the image, and \acrshort{geminipro} can correctly identify some of them, such as ``background''. However, its reflection fails to initiate a corrective action.

Moreover, the \acrshort{cot} of the worst result (Fig.~\ref{fig:geminipro-worst-coco}) reveals a fundamental flaw in \acrshort{geminipro}'s visual perception module and visual reasoning process. It mistakenly identifies the entire picture as ``net'', while in other parallel attempts, the vision system can identify details like fences, which implies that its vision system is highly unstable with inconsistent performance. Furthermore, its multimodal \gls{cot} reasoning fails to correct its mistakes due to the issue of pretended reflections. During its examination of the generated mask, it concludes that:
\begin{quote}
    \textit{I've examined the segmentation mask to ensure the color values align with the expected category. The 'net' category is properly represented by magenta, and this color fills the entire mask as required. It's a precise mapping of the visual element. I'll focus on the next step.}
\end{quote}

This extreme example hints at further investigation. One cause may be that its internal reasoning module fails to incorporate the visual information of the image and its generated mask into its reasoning and reflection process while it learns the superficial form of reflection during training.

\begin{figure}
    \centering
    \includegraphics[width=1\linewidth]{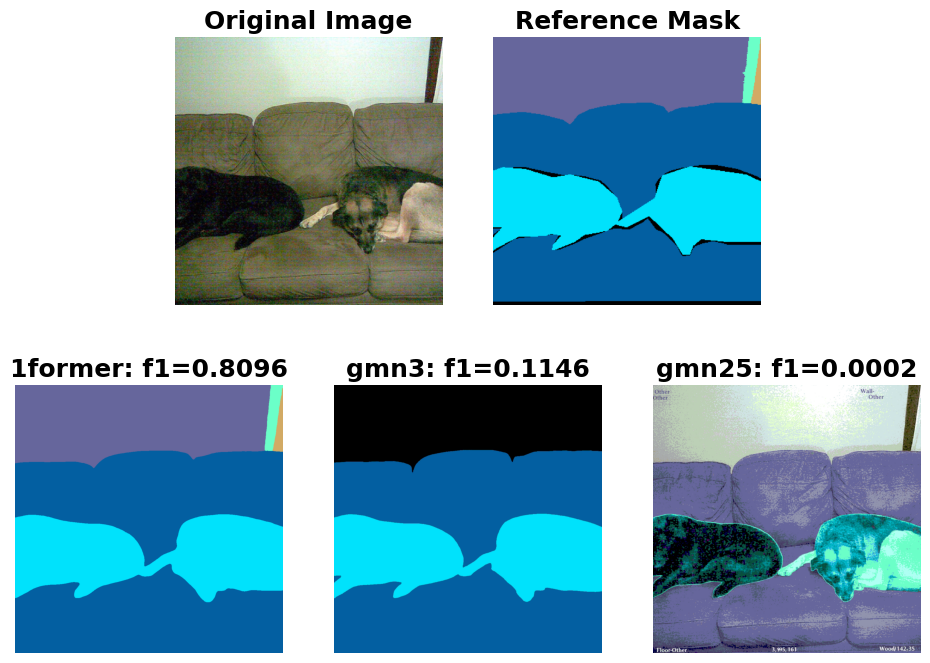}
    \caption{Comparison between the Results of Different Models on \coco{}. Their short code names and F1 scores are on the top of the images. We picked the image on which \acrshort{oneformer} achieved the highest F1 score $0.8096$ while \acrshort{geminipro} achieved $0.1146$.}
    \label{fig:model-comparison-coco}
\end{figure}

\begin{figure}
    \centering
    \includegraphics[width=1\linewidth]{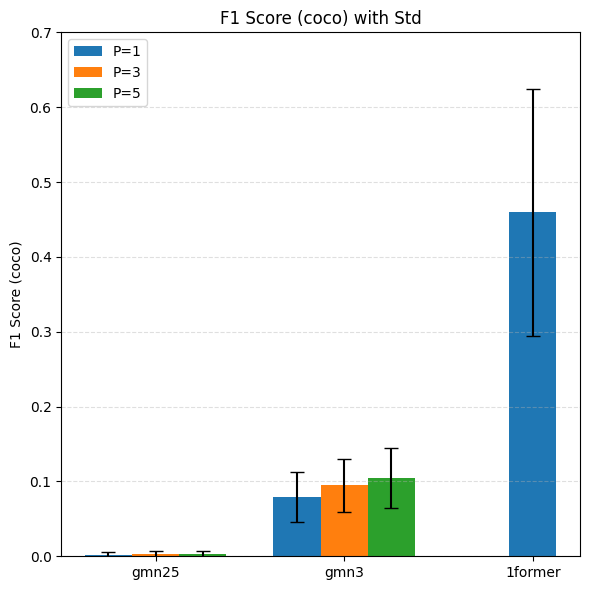}
    \caption{F1 Scores for experiments on \coco{}. For $p=[3, 5]$, we ask \glspl{omm} to try 3 or 5 times and select the best result in these attempts. As \acrshort{oneformer} contain no stochastic components, we did not run more attempts. Due to their poor performance in previous experiments, we did not run experiments for other \glspl{omm}.}
    \label{fig:f1-comparison-coco}
\end{figure}

Despite the extreme failures, the outcomes of multiple attempts from \acrshort{geminipro} are still plausibly good, in contrast to the results of \acrshort{gemini}, as shown in Fig.~\ref{fig:model-comparison-coco}, where \acrshort{gemini} completely failed to generate an \gls{ss} mask. Our quantitative results in Fig.~\ref{fig:f1-comparison-coco} align with our qualitative analysis as well. For mIoU and Dice, please refer to Figs.~5 and~6 in Suppl.~C.

\section{Limitations and Discussion}

We benchmark recent \glspl{omm} on the subsets of CelebAMask-HQ~\cite{CelebAMask-HQ} and COCO~\cite{coco-dataset} in terms of F1, mIoU, and Dice scores. We further analyse the results and discover the limitations of existing \glspl{omm}. Here, we discuss our limitations and future directions in terms of data, models, and metrics.

\textbf{Datasets:} Due to resource constraints, we did not conduct the experiments on the entire CelebAMask-HQ~\cite{CelebAMask-HQ} and COCO~\cite{coco-dataset} datasets. The quantitative results may be biased towards these subsets. We did not experiment with the models on more segmentation datasets (\eg, SA-CO~\cite{sam3}). However, \pa{} can be easily extended to other segmentation datasets. Therefore, we leave this as future work.

\textbf{Data Refinement and Dataset Development:} Because many of the results provided by \glspl{omm} (\eg, results by \gls{geminipro} in Figs.~\ref{fig:model-compare},~\ref{fig:best-gemini-pro}, and~\ref{fig:segface-gemini}) are good enough to serve as initial drafts for human annotation or even better than human annotations, in the future, we can use \glspl{omm} to re-examine the data and labels in existing segmentation benchmarks and refine their data quality, as well as improve the efficiency of annotation in new benchmarks.

\textbf{Model Selection:} We do not benchmark against RAS~\cite{ras}, SAM Agent~\cite{sam3}, or SAM4MLLM~\cite{sam4mllm}. The performance difference between native generation (\ie, ours) and these model integration methods remains to be seen. However, similar to specialized models (\eg, \acrshort{segface} and \acrshort{oneformer}), these methods have unfair advantages over \glspl{omm} since they are designed and trained for specific \gls{ss} tasks. Therefore, similar to the results generated by the specialized models, the results from RAS~\cite{ras}, SAM Agent~\cite{sam3}, and SAM4MLLM~\cite{sam4mllm} are not comparable to those from \glspl{omm}.

\textbf{Better Prompts for \glspl{omm}:} As seen in Figs.~\ref{fig:best-gemini-pro} and ~\ref{fig:worst-gemini-pro}, one subtle but noticeable difference is in the eye areas. In reference masks, the masks for the eyes cover only the eyeballs, but in the predictions of \acrshort{geminipro}, the masks cover most of the periorbital regions. Such a difference may be due to our under-specification of the task, as we did not mention in our prompt whether it should label only the eyeballs. If we provide more detailed instructions, we may be able to improve performance further. 

\textbf{\gls{omm} Research:} The reason why the shuffled color encodings improve the performance of \acrshort{geminipro} remains an interesting topic that may be closely related to its vision system and visual reasoning capabilities. Furthermore, if we could gain access to the source code and weights of \gls{geminipro}, the examples of fake quality control may be valuable for mechanistic interpretability research~\cite{mechanisticinterpretabilityaisafety} to recover the mechanisms of its internal visual and generative systems.

\textbf{Metric Design:} As we have seen in Fig.~\ref{fig:segface-gemini}, the score discrepancy between \acrshort{segface} and \acrshort{geminipro} is significant; however, it does not reflect the visual similarity between the two masks. As \glspl{omm} may be significantly valuable in many applications (\eg, refining the data of existing \gls{ss} datasets), we need metrics that are better than F1 Score and mIoU to evaluate their performance and guide related research.
\section{Conclusion}

We present \pa{}, in which we propose using segmentation tasks to probe the pixel-precision visual intelligence of advanced \glspl{omm}. We use semantic segmentation tasks on CelebAMask-HQ~\cite{CelebAMask-HQ} and COCO~\cite{coco-dataset} to test the \gls{ppvi} of frontier \glspl{omm} (\ie,~\cite{gemini3proimage,gemini25flashimage,gpt4o,emu35,unimoe2}). With our benchmark, we find that \acrlong{geminipro} represents a major breakthrough in this front. With qualitative and quantitative results, it demonstrates superior performance under our \textit{zero-shot setting}. We also present failure cases of these models and discuss their failure modes, which shed light on potential future research directions in dataset development, \gls{omm} research, and metric design.

\clearpage
\bibliographystyle{named}
\bibliography{ijcai26}

% Supplementary material is compiled separately in ijcai26_suppl.tex

\end{document}